\title{Learning Multivariate CDFs and Copulas using Tensor Factorization}

\author{Magda Amiridi,  Nicholas D. Sidiropoulos
\\ University of Virginia
   \\   \{ma7bx,nikos\} @virginia.edu} 

\newcommand{\abstractText}{\noindent
Learning the multivariate distribution of data is a core challenge in statistics and machine learning. Traditional methods aim for the probability density function (PDF) and are limited by the curse of dimensionality. Modern neural methods are mostly based on black-box models, lacking identifiability guarantees. In this work, we aim to learn multivariate cumulative distribution functions (CDFs), as they can handle mixed random variables, allow efficient `box' probability evaluation, and have the potential to overcome local sample scarcity owing to their cumulative nature. We show that {\em any} grid-sampled version of a joint CDF of mixed random variables admits a {\em universal} representation as a naive Bayes model via the Canonical Polyadic (tensor-rank) decomposition. By introducing a low-rank model, either directly in the raw data domain, or indirectly in a transformed (Copula) domain, the resulting model affords efficient sampling, closed form inference and uncertainty quantification, and comes with uniqueness guarantees under relatively mild conditions. We demonstrate the superior performance of the proposed model in several synthetic and real datasets and applications including regression, sampling and data imputation. Interestingly, our experiments with real data show that it is possible to obtain better density/mass estimates indirectly via a low-rank CDF model, than a low-rank PDF/PMF model.
}

\documentclass[paper=a4, fontsize=12pt,
  DIV=10, twocolumn]{article}
\usepackage{enumitem}%
\usepackage{bbold}
\usepackage{amsmath,amssymb,color,enumerate}
\usepackage{graphicx}
\usepackage{calligra}
\usepackage{mathtools}
\usepackage{algorithm}
\usepackage{tabstackengine}

\usepackage{wrapfig}
\usepackage{algcompatible}

\usepackage{bbm}
\usepackage{booktabs}
\usepackage{graphicx} 
\usepackage{cite}
\usepackage{amsmath,amssymb,amsfonts}
\usepackage{algorithm}

\usepackage{subfig}
\usepackage{xcolor}
\usepackage[hidelinks]{hyperref}

\newtheorem{Lemma}{Lemma}

\newcommand{\vtheta}{{\boldsymbol\theta}}

\usepackage{xurl}
\usepackage[super,comma,sort&compress]{natbib}
\usepackage{abstract}

\usepackage{hyperref}
\hypersetup{colorlinks=true, urlcolor=blue, linkcolor=blue, citecolor=blue}

\begin{document}


\twocolumn[
  \begin{@twocolumnfalse}
    \maketitle
    \begin{abstract}
      \abstractText
      \newline
      \newline
    \end{abstract}
  \end{@twocolumnfalse}
]

\section{Introduction} 
Modeling complex data distributions is a task of central interest in statistics and machine learning. Given an accurate and tractable  estimate of the joint distribution function, various kinds of statistical tasks can follow naturally including fast sampling, tractable computation of expectations, and deriving conditional and marginal densities. To list a few recent applications, such models have demonstrated success in  generating high-fidelity images \cite{karras2019style}, \cite{brock2018large}, realistic speech synthesis \cite{hsu2017learning}, \cite{oord2016wavenet}; semi-supervised learning \cite{odena2016semi}; reinforcement
learning \cite{ostrovski2017count}; and detecting adversarial data \cite{mohaghegh2020advflow}. The purpose of this work is to introduce a new class of universal estimators for multivariate distributions based on CDFs and the Canonical Polyadic (tensor-rank) decomposition, and to demonstrate their direct applicability and efficiency in missing data imputation, sampling, density estimation, and regression tasks. 

 Distribution modeling is often studied under the perspective of non-parametric PDF estimation, in which histogram, kernel \cite{rosenblatt1956}, \cite{parzen1962estimation}, and orthogonal series methods \cite{girolami2002orthogonal}, \cite{efromovich2010orthogonal}, \cite{tsybakov2008introduction} are popular approaches with well-understood statistical properties. Although these estimators are data-driven and do not impose restrictive parameterisations on the form of the data distribution, they usually have poor performance on datasets of high dimensions because of the ``curse of dimensionality''. Currently, the most prominent methods for modeling multivariate distributions rely on neural networks \cite{dinh2016density}, \cite{papamakarios2017masked},    \cite{kingma2013auto}, \cite{goodfellow2014generative}. Such methods are capable of modeling higher dimensional data such as images and sound, either implicitly or explicitly. However, most of them are black-box models \cite{chilinski2020neural} without any identifiability guarantees, lacking the simplicity and interpretability of the classical methods. Additionally, they lack the ability to efficiently compute expectations, marginalize over subsets of variables, and evaluate conditionals, which is limiting in many critical machine learning applications.

 This paper attempts to bridge the gap between principled traditional non-parametric statistics and the scalability benefits of neural-based models by developing two variants of a rank-constrained estimator for multivariate CDFs based on tensor rank decomposition – known as Canonical Polyadic (CP) or CANDECOMP/PARAFAC Decomposition  \cite{hitchcock1927expression}, \cite{Harshman1970}, \cite{sidiropoulos2017tensor}. Our starting point is that any grid-sampled version of an $N$-dimensional CDF is an $N-$way cumulative probability tensor $\widehat{\mathcal{F}}$, evaluated on a predefined ${I_1 \times I_2 \times \cdots \times I_N}$ grid $\mathit{G}$.  We will refer to $\widehat{\mathcal{F}}$ as grid-sampled CDF tensor. The evaluation grid $\mathit{G}$ describes the (finite) levels/cut-offs of the CDF for every dimension and can be taken to be the cartesian product of the training samples in each dimension, or reduced via scalar or vector k-means. Each element of $\widehat{\mathcal{F}}$ can be easily estimated via sample averaging from realizations of the random vector of interest. 

Any tensor can be decomposed as a sum of $R$ rank-$1$ tensors, for high-enough but finite $R$ \cite{sidiropoulos2017tensor}. To maintain direct control over the number of tensor parameters with growing dimensionality $N$, which entails $\mathcal{O}\left(\prod_{n=1}^N I_n\right)$ CDF tensor elements, we introduce the reconstructed approximation of $\widehat{\mathcal{F}}$, using the rank-$R$ ${N}$-dimensional parameterization ${\mathcal{F}} \in \mathbb{R}^{{I_1 \times I_2 \times \cdots \times I_N}}$. Such parametrization has much fewer degrees of freedom designated by the rank and size of the the tensor. Tensor decompositions arise as a powerful tool for extracting meaningful latent structure from given data  and can  encode the salient characteristics of the multivariate grid-sampled CDF tensor $\widehat{\mathcal{F}}$. We seek to minimize the squared loss between $\mathcal{F}$ and the empirical CDF tensor $\widehat{\mathcal{F}}$. We formulate this task both directly, i.e., by forming and decomposing $\widehat{\mathcal{F}}$, and indirectly, as a hidden tensor factorization problem, i.e., the pertinent parts of the latent factors of the CPD model are updated from rank-1 measurements of the ``hidden'' discretized CDF tensor $\widehat{\mathcal{F}}$. From an algorithmic perspective, we propose alternating optimization, where each matrix factor is updated using ADMM, as well as stochastic optimization using Adam to allow scaling to larger datasets. 

\subsection{Contributions}  In summary, the present paper shows that:
\begin{itemize}[noitemsep,topsep=0pt]
    \item {\em Every} multivariate CDF evaluated on a predefined grid
admits a compact representation via a latent variable naive Bayes model with bounded number of hidden states equal to the rank of the grid-sampled CDF tensor. 
\item  This affords easy sampling, marginalization (by discarding the subset of factor matrices corresponding to the variables we are not interesting in), derivation of conditional distributions and expectations, and uncertainty quantification -- bypassing the curse of dimensionality.   
\item The proposed model also affords direct and efficient estimates of (possibly semi-infinite) ``box'' probabilities, which is important for classification tasks. Multivariate PDF estimators, on the other hand, require multidimensional integration (analytical, numerical, or sampling-based Monte-Carlo) to estimate box probabilities, which is cumbersome and often intractable.
\item  At the same time, the proposed rank-constrained estimator is unassuming of the structure of the data (thus offering  greater expressive power) and identifiable under relatively mild rank conditions -- see~\cite{sidiropoulos2017tensor}. 
\item  On the experimental side, our results indicate that,  perhaps surprisingly, {\em estimating the grid-sampled CDF and then deriving a PDF estimate from it yields improved performance relative to direct PDF estimation in several machine learning applications of interest}. In addition, the performance of {\em the proposed non-parametric model in the Copula domain outperforms state-of-the-art Copula-based baselines}.
\end{itemize}

\section{Background}
\subsection{Related work} 
Unsupervised learning of multivariate distributions has seen tremendous progress over recent years, for the case of PDF modeling in particular. Classical methods in the literature include kernel density estimation (KDE)~\cite{rosenblatt1956, parzen1962estimation}, histogram density estimation (HDE), and Orthogonal Series Density Estimation (OSDE) \cite{girolami2002orthogonal, efromovich2010orthogonal, tsybakov2008introduction}. All  of the aforementioned methods, however are inefficient for datasets with higher dimensionalities. Neural network-based approaches for distribution estimation have recently shown promising results in  high-dimensional problems. Auto-regressive (AR) models such as \cite{salimans2017pixelcnn, oord2016wavenet}  decompose the distribution into a product of conditionals, where each
conditional is modeled by a parametric distribution (e.g., Gaussian or mixture of Gaussians in the continuous case). Normalizing flows (NFs) \cite{papamakarios2021normalizing} represent a density value though an invertible transformation
of latent variables with known density.  

On the down-side, AR models are naturally sensitive to the order of the variables/features while strong network constraints of NFs can be restrictive for model expressiveness. Most importantly, AR and NF do not yield an explicit estimate of the density function; they are `oracles' that can be queried to output {\em an estimate of the density at any given input point}, i.e., to generate samples of the sought density -- the difference is important. Therefore, given a trained model, calculating expectations, marginal and conditional distributions is not straightforward with these methods. The same holds for generative adversarial networks \cite{goodfellow2014generative} (GANs) as they do not allow for likelihood evaluation on held-out data. Furthermore, deep multivariate CDF based models such as \cite{chilinski2020neural}, do not address model identifiability, and can not guarantee the recovery of the true latent factors that generated the observed samples.

\textbf{Tensor modeling of distributions}: Tensor models for estimating distributions have been proposed for both discrete and  continuous variables. In the discrete case, the work in \cite{kargas2018tensors} showed that any joint PMF can be represented as an $N$-way probability tensor and by introducing a CPD model, every multivariate PMF can be represented by a latent variable naive Bayes model with a finite number of latent states. For continuous random vectors, the joint PDF can no longer be directly represented by a tensor. Earlier work (\cite{song2014nonparametric}) has dealt with latent variable models, but not general distributions. In contrast to prior work (\cite{song2014nonparametric}, \cite{kargas2019learning}), we make no assumptions regarding a multivariate mixture model of non-parametric product distributions in this paper. Another line of work (see \cite{amiridi2020nonparametric}, \cite{amiridi2021low}) proposed a ``universal'' approach for smooth, compactly supported multivariate densities by representing the underlying density in terms of a finite tensor of leading Fourier coefficients.  Our work requires less restrictive assumptions, as it also works with discrete or mixed random variables, of possibly unbounded support.
\subsection{Notation, Definitions, and Preliminaries}
We use the symbols $\mathbf{x}$, $\mathbf{X}$, $\mathcal{X}$ for vectors, matrices and tensors respectively. We use the notation $\mathbf{x}(n)$, $\mathbf{X}(:,n)$, $\mathcal{X}(:,:,n)$ to refer to a particular element of a vector, a column of a matrix and a slab of a tensor.  Symbols $\circ$,  $\otimes$, $\circledast$, $\odot$  denote the outer, Kronecker, Hadamard and Khatri-Rao (column-wise Kronecker) product respectively. The vectorization operator is denoted as ${\rm vec}(\mathbf{X})$, ${\rm vec}(\mathcal{X})$ for a matrix and tensor respectively \cite{sidiropoulos2017tensor}. Additionally, ${\rm diag}(\mathbf{x})\in \mathbb{R}^{I \times I}$ denotes the diagonal matrix with the elements of vector $\mathbf{x}\in \mathbb{R}^I$ on its diagonal. Symbols $\|\mathbf{x}\|_1$, $\|\mathbf{x}\|_2$, $\|\mathbf{X}\|_F$, and $d_{TV}$ correspond to $L_1$ norm, $L_2$ norm, Frobenius norm, and total variation distance. 
The {total variation distance} between distributions
$p$ and $q$ is defined as $d_{TV}(p,q) = \frac12\|p - q\|_1.$

Given an $N$-dimensional random vector  ${\mathbf{X}} := \left[X_1,\ldots, X_N\right]^T$, ${{\mathbf{X}}} \sim {F}_{{\mathbf{X}}}$ will denote that the random vector ${\mathbf{X}} $ follows distribution ${F}_{{\mathbf{X}}}$. $\mathbf{1}(A)$ is the indicator function of event $A$, i.e., it is 1 if and only if $A$ is true. The set of integers $\{1,\ldots, N\}$ is denoted as $[N]$. Given $M$ data samples, $\mathcal{D} =  \left\{ {\mathbf{x}}_m  \right\}_{m=1}^M$ denotes the given dataset.

\textbf{Tensor Preliminaries:}
An $N$-way tensor ${{\mathcal{F}} \in \mathbb{R}^{I_1 \times I_2 \times \cdots \times I_N}}$ is a multidimensional array whose elements are a function of $N$ indices $\mathcal{F}(i_1,i_2,\cdots, i_N)$, %
with $i_n$ ranging from $1,\cdots,I_n$. Similar to the matrix case, a tensor can be represented in succinct form with tensor decompositions. It can always be decomposed as a finite sum of $R$ rank-$1$ tensors for high-enough but finite $R$, i.e., $
\mathcal{F} = \sum_{h=1}^R{ \boldsymbol{\lambda}}(h) \mathbf{A}_1(:,h) \circ \mathbf{A}_2(:,h), \cdots, \mathbf{A}_N(:,h),$
where $\boldsymbol{\lambda} \in {\mathbb{R}}^{R}$, each $\mathbf {A}_n \in {\mathbb{R}}^{I_{n}\times R}$, ${\bf A}_n(:,h)$ denotes the $h$-th column of matrix ${\bf A}_n$. The decomposition can be compactly denoted as a collection of latent factor matrices and the weight vector $\boldsymbol{\lambda}$, $\mathcal{F}= [\![\boldsymbol{\lambda}, \mathbf{A}_1,\mathbf{A}_2, \cdots,\mathbf{A}_N]\!]$ or elementwise as
$\mathcal{F}(i_1,i_2,\cdots, i_N) =\sum_{h=1}^R\boldsymbol{\lambda}(h)\prod_{n=1}^N \mathbf{A}_n(i_n,h).$
When $R$ is minimal, it is called the rank of ${\mathcal{F}}$, and the decomposition is called Canonical Polyadic Decomposition (CPD) \cite{hitchcock1927expression,Harshman1970}. For many practical applications, low-rank decompositions can be used for extracting latent factors from tensorial datasets. 

If the tensor can be well approximated by a low-rank CPD model, the number of free parameters of the model to be estimated drops to ${O}\big(R\left(\sum_{n=1}^N I_n\right)\big)$. A key property of the CPD  is that the rank-1 components are unique under mild conditions. For a rank$-R$ tensor, the goal is to recover all the underlying factors $\mathbf{A}_n$ and $\boldsymbol{\lambda}$. The uniqueness properties of tensor rank decomposition \cite{sidiropoulos2000uniqueness} implies probability (and generative) model identifiability in our context. 

\section{Methodology}
\subsection{Sketching multivariate CDFs}

Let ${\mathbf{X}} \in\mathbf{R}^N$, ${{\mathbf{X}}} \sim {F}_{{\mathbf{X}}}$, be a random vector comprising $N$ discrete/categorical or continuous constituent random variables (features) $\mathbf{X}:= \left[X_1,\ldots, X_N\right]^T$. Given a collection $\mathcal{D} =  \left\{ \mathbf{x}_m  \right\}_{m=1}^M$ of independently and identically generated observations $ \mathbf{x}_m =[\mathbf{x}_m (1),\ldots, \mathbf{x}_m (N)]^T \in\mathbf{R}^N$ sampled from ${F}_{{\mathbf{X}}}$, the goal is to find an accurate estimate ${\widehat{F}}_{{\mathbf{X}}}$ of the true distribution function ${F}_{{\mathbf{X}}}$ with limited assumptions about the underlying model. 


Let us denote the CDF cut-offs of a variable $X_n$ as  $[ x_{1}(n), \ldots,  x_{I_n}(n)]^T\in\mathbf{R}^{I_n}$. In each dimension $n\in[N]$, the cut-off points are sorted in increasing order, e.g.,  $x_{1}(n)<\cdots<x_{i_n}(n)<\cdots< x_{I_n}(n)$.  Considering an evaluation (target) point ${\mathbf{x}}_{\boldsymbol{i}}=[x_{i_1}(1), x_{i_2}(2),\ldots, x_{i_N}(N)]^T$, a natural estimator of the sought CDF is the empirical cumulative distribution function (ECDF) ${\widehat{F}}_{{{\mathbf{X}}}}$, 
$
{\widehat{F}}_{{\mathbf{X}}}({\mathbf{x}}_{\boldsymbol{i}})=
\frac{1}{M}\sum_{m=1}^M \mathbf{1}\left({{\mathbf{x}}}_m\leq{\mathbf{x}}_{\boldsymbol{i}}\right)=\frac{1}{M}\sum_{m=1}^M\mathbf{1}\left( \mathbf{x}_m (1)\leq x_{i_1}(1),\ldots, \mathbf{x}_m (N)\leq x_{i_N}(N)\right)$.

By considering an ${I_1 \times I_2 \times \cdots \times I_N}$  grid $\mathit{G}$ (with possibly non-uniform mesh) of target evaluation points ${\mathbf{x}}_{\boldsymbol{i}}$, defined as the cartesian product of the observed 1-D samples (or a reduced set of surrogates obtained via k-means), one can obtain a discretized version of the ECDF, which we will be referring to as grid-sampled (empirical) CDF. We denote the multivariate CDF evaluation grid $\mathit{G}$ as $\mathit{G} =\{\left[x_{i_1}(1), x_{i_2}(2),\ldots, x_{i_N}(N)\right]^T\in R^N \text{ }|\text{ }i_1 \in [I_1], i_2 \in [I_2], \ldots, i_N \in [I_N]\}$. 

\textbf{Motivation:} Approximating the cumulative distribution function, instead of the multivariate density \cite{kargas2019learning}, has significant advantages. A joint CDF (and its grid-sampled sketch) always exists in contrast to a PDF/PMF. Instead of considering one type of random variables, a CDF can simultaneously model discrete, continuous or mixed random variables. All observations ``up to'' a point ${\mathbf{x}}_{\boldsymbol{i}}$ are considered for a single point CDF estimate  $F_{{\mathbf{ X}}}({\mathbf{x}}_{\boldsymbol{i}})$, allowing estimation in places where there is a scarcity of ``local'' samples. Additionally, it  affords direct and efficient estimates of (possibly semi-infinite) ``box'' probabilities, e.g. of type ${Pr}\{35 < \text{Age} \leq 45, \cdots, 55 <  \text{Income} \leq 70\}$, avoiding the (potentially intractable) multi-dimensional integration of the joint PDF. These probabilities are needed for classification and anomaly detection. For example, in the $3$-D case, only $8$ basic operations (additions, subtractions) of the multivariate CDF are required for estimating a single interval probability, whereas a $3$-D integration is required in the joint PDF case, and the problem is compounded in higher-dimensional cases, where the only tractable option is Monte-Carlo integration, which has poor accuracy as it suffers from the curse of dimensionality -- an exponential number of Monte-Carlo samples is needed for accurate estimation. 

Still, sufficiently reliable estimates of ${\widehat{\mathcal{F}}}_{{{\mathbf{X}}}}$ in high-dimensional spaces require a number of observations that grows quickly with the tensor order $N$, $\mathcal{O}\left(\prod_{n=1}^N I_n\right)$. To reduce the number of free parameters and scale up to higher dimensions, certain assumptions must be made. In our case, the sole assumption is that the CDF tensor can be well-approximated by a low-rank model ${{\mathcal{F}}}_{{\mathbf{X}}}$. CPD is a powerful model that can parsimoniously represent the high-order interactions among multi-way data exactly or approximately, leading to significant reduction in the number of parameters. Even in high dimensional settings, data can be approximated to live in relatively low dimensional space by exploiting the low rank property ~\cite{kolda2009tensor}, \cite{sidiropoulos2017tensor}. 
 
\subsection{Modeling and Interpretation}
So far we have shown that given a set of $M$ realizations in the training set and an evaluation grid $\mathit{G}$ indexed by $[i_1, i_2,\ldots,i_N]^T$, the multivariate grid-sampled empirical CDF induced by $\mathit{G}$ can always be represented as a finite CDF tensor ${\widehat{\mathcal{F}}}_{{{\mathbf{X}}}}$, whose elements can be estimated by the following sum \begin{equation}
{\widehat{\mathcal{F}}}_{\mathbf{X}}(i_1, i_2,\ldots,i_N)=\frac{1}{M}\sum_{m=1}^M \mathbf{1}\left({{\mathbf{x}}}_m\leq{\mathbf{x}}_{\boldsymbol{i}}\right).
\label{empirical}
\end{equation}
Assuming that $\{\mathbf{x}_{m}\}_{m=1}^M$ is i.i.d. in $m$,
it is easy to verify that $\widehat{\mathcal{F}}$ is an unbiased estimator of the corresponding joint cumulative distribution function \cite{Shao_2003_book}, i.e.,$
\mathbb{E}\left[ {\widehat{\mathcal{F}}}_{{{\mathbf{X}}}}(i_1, i_2,\ldots,i_N) \right] = {{F}}\left(x_{i_1}(1), \ldots, x_{i_N}(n)\right),$
 for all $\left[x_{i_1}(1), \ldots, x_{i_N}(n)\right]^T\in \mathit{G}$ and $M \geq 1.$ 
 
\begin{figure*}
\center
\includegraphics[width = 1.2 \columnwidth]{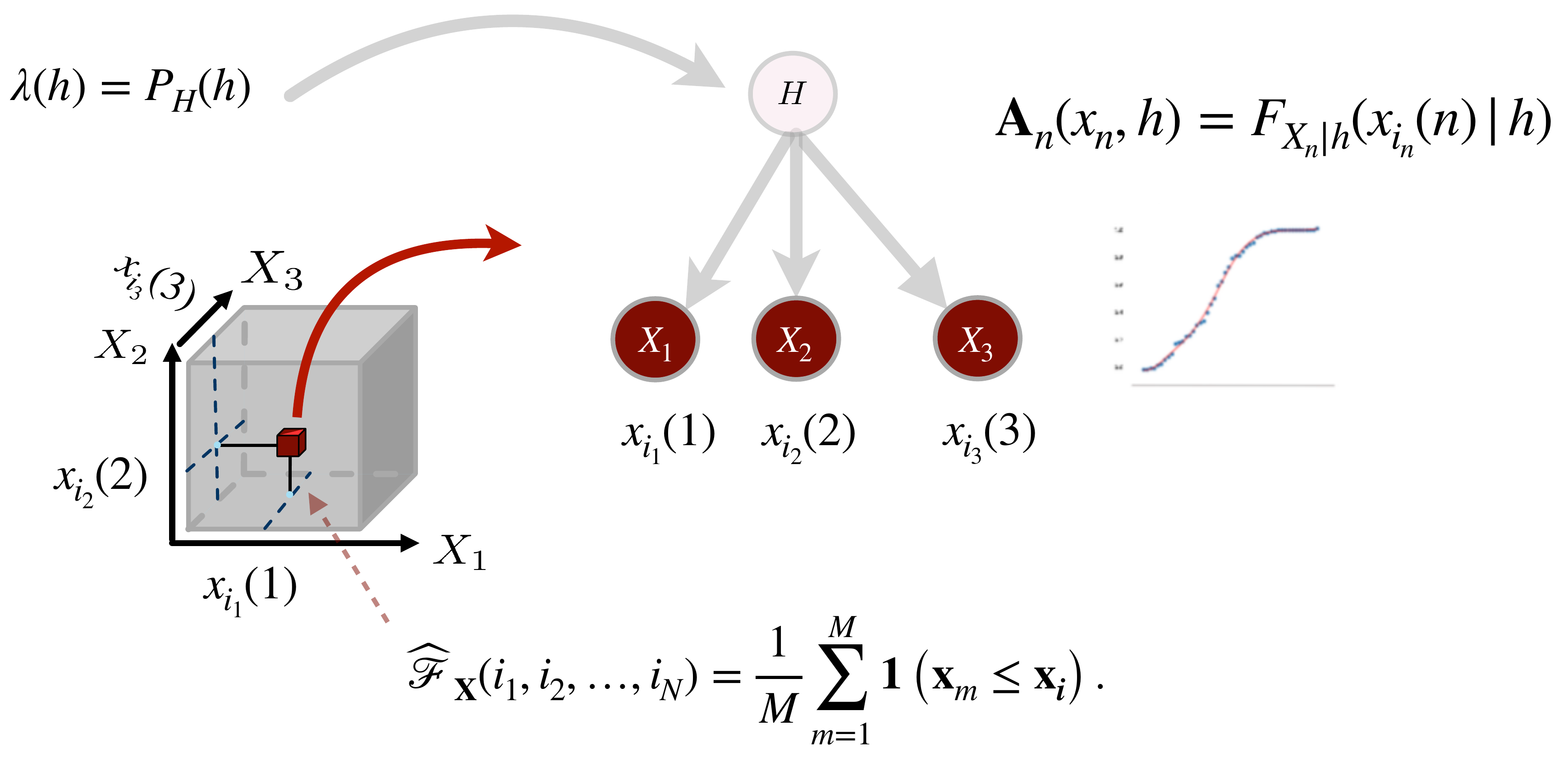}
\caption{Samples from the empirical grid-sampled CDF tensor can be thought as being generated via a latent variable naive Bayes model. This generative model says that observed data samples can be thought to be generated by a two-step process: First a value for $H$ is generated according to $P_H(h)$ and then, a value $x_{i_n}(n)$ is generated according to ${F}_{X_n|h}$.}
\label{fig:NB}
\end{figure*}

We will leverage the universal approximation ability of tensors by a CPD model to build a parameterization of multivariate grid-sampled  (empirical) CDFs. We focus on the first $R$ {\em principal components} of the resulting tensor, i.e., we introduce a low-rank parametrization  ${{\mathcal{F}}}_{{\mathbf{X}}}$ of the grid-sampled {\em CDF tensor} ${\widehat{\mathcal{F}}}_{{{\mathbf{X}}}}$, ${{\mathcal{F}}}_{{\mathbf{X}}}(i_1, i_2,\ldots,i_N)
 = \sum_{h=1}^R \boldsymbol{\lambda}(h) \prod_{n=1}^{N} \mathbf{A}_n(i_n,h)\label{eq:1}$. By enforcing non-negativity and ``valid CDF'' constraints on each column of the CPD factors, i.e., $\mathbf{A}_n(,h)$ to be non-negative, non-decreasing, and the last element to be equal to $1$, as well as simplex constraints ${\mathbf{1}}^T \boldsymbol{\lambda} = 1$ on $\boldsymbol{\lambda}$, it can be shown, using an argument first made for the case of multivariate PMFs by \cite{kargas2018tensors}, that every multivariate grid-sampled CDF can be thought to be generated according to a hidden variable model,
\begin{equation}
{{\mathcal{F}}}_{{\mathbf{X}}}(i_1,\ldots,i_N)=\sum_{h=1}^R P_H(h)  \prod_{n=1}^{N} {F}_{X_n | H} (x_{i_n}(n)| h).\label{eq:2}
\end{equation}
This hidden variable $H$ takes $R$ possible values (R is the tensor rank) and follows a prior distribution ${P}_H( h)={Pr}(H = h)$. According to the resulting model, given $H=h$, the manifest variables  $X_1,\ldots, X_N$ are generated independently via an unknown mapping expressed by $N$ univariate (discretized) conditional distributions ${F}_{X_n|H}(x_{i_n}(n)|h)={Pr}(X_n \leq x_{i_n}(n)|H=h)$. This model is known as latent variable naive Bayes (NB) model and a visualization is presented in Figure \ref{fig:NB}. 

Remarkably, such representation is {\em universal} if one allows the hidden variable to have a sufficiently rich finite alphabet. 
The parameters of this model $F_{X_n|H}, {P}_H$ will be recovered (or estimated) jointly by decomposing tensor ${\widehat{\mathcal{F}}}_{{{\mathbf{X}}}}$. A key property of the CPD  is that the rank-$1$ components are unique under mild conditions. Uniqueness offers identification guarantees under ideal conditions -- that is, when the true model is low rank and the correct rank is used. For model approximation, uniqueness means that there is a single model of the data {\em for fixed residuals} that is consistent with what we have observed. See~\cite{sidiropoulos2017tensor} for detailed identifiability results. 

The result in (\ref{eq:2}) states that a decomposition of  multivariate grid-sampled CDFs in terms of $1-D$ conditionals can be computed via constrained CPD. Conversely, if one {\em assumes} that a multivariate CDF is a finite mixture of $R$ product distributions of mixed random variables, the discretized conditional CDFs are identifiable and can be recovered by decomposing the exact CDF tensor,
\begin{align*}
{{F}}_{{\mathbf{X}}}{({{\mathbf{x_i}}})}&=\mathbb{E}\left[\mathbf{1}\left( X_1\leq x_{i_1}(1),\ldots, X_N\leq x_{i_N}(N)\right)\right]\nonumber\\
&=\sum_{h=1}^R{P}_H(h)\prod_{n=1}^{N}\mathbb{E}\left[\mathbf{1}{\{X_n\leq x_{i_n}(n)\}}|h\right]\nonumber\\
&=\sum_{h=1}^R{P}_H(h)\prod_{n=1}^{N}{F}_{X_n|h}(x_{i_n}(n)|h)\nonumber.
\end{align*}

Let us consider an evaluation grid of size $I^N$ and refer to the element-wise grid spacing  as `resolution'. Using Lemma C.1. of \cite{daskalakis2012learning}, we have obtained the following sample complexity result for our problem. A detailed proof of the Lemma below can be found in the supplementary material Section~\ref{sample_complx}. The result in Lemma \ref{lem:sample_complexity} states that it is possible to learn, within  error $6{\epsilon'}$, any rank$-R$ grid-sampled distribution with resolution $\epsilon'$, using a number of samples that is (worst-case) log-linear in rank $R$ and quadratic in the number of dimensions $N$.

\begin{Lemma}
\label{lem:sample_complexity}
Let $F$ be any unknown $N-$ dimensional (``true'') grid sampled CDF of ${\mathbf{X}}$ with rank $R$, alphabet size $I$ (per variable), and resolution parameter $\epsilon'$. There is an algorithm that uses 
\resizebox{.9\linewidth}{!}{\
  \begin{minipage}{\linewidth}
  \begin{equation*}
\mathcal{O}\left( {\epsilon'}^{-2}({RNI\text{log}\frac{1}{\epsilon'}+NI R\text{log} R+RN^2I
\text{log}(2I)}) \cdot \text{log}(\frac{1}{\delta})\right)
\end{equation*}
  \end{minipage}
}

samples from $F$, and with probability $1-\delta$ outputs a distribution $F'$ that satisfies $d_{TV}(F,F')\leq 6{\epsilon'}.$
\end{Lemma}

Thanks to the naive Bayes model interpretation, we can go from the joint CDF to the joint PMF/PDF domain efficiently, by taking $1$-D differences / derivatives. In the case of continuous variables (PDF), we first use  interpolation (linear, bandlimited, or spline -- we use linear for simplicity) followed by $1$-D differentiation. For PDFs that are (approximately) band-limited with cutoff frequency $\omega_c$, \cite{kargas2019learning} showed that they can be recovered from uniform samples of the associated CDF taken $\frac{\pi}{\omega_c}$ apart. In this way, one is able to easily extract any conditional or marginal distribution at negligible complexity beyond what is needed to estimate the grid-sampled CDF tensor factorization --  we simply drop factor matrices corresponding to the variables we are not interested in. We stress that multivariate interpolation/differentiation is not needed to compute any PDF, due to the separable nature of the CPD model. 

Thus, given a test sample it is easy to infer a response value or label. Let us assume that the label information is stored in the $N-$th variable. Given possibly incomplete data (vector realizations with missing entries) specified by the set $X_S, S\subseteq V$, with $X_V=\{X_1,X_2,\ldots,X_{N-1}\}$ , we can easily compute the posterior probability $P_{X_N| X_S}$
and derive any desired classifier or estimator and estimate its uncertainty (e.g., conditional variance, conditional tail probability) at the same time. Such uncertainty quantification is very useful in many applications.

\textbf{Preprocessing step:} We explore two different variants. The first method, LR-CDF, directly fits a low-rank tensor model of the grid-sampled CDF tensor, given raw data samples ${{\mathbf{x}}}_{m} \in \mathbb{R}^N, m\in[M]$, without preprocessing. The second approach, LR-Copula, introduces a low-rank parametrizaton of the grid-sampled version of the Copula instead. A Copula is a special family of CDFs with uniform marginals and it is invariant under nonlinear monotonic transformations of the individual variables. For LR-Copula, before training the models, we map each sample vector ${{\mathbf{x}}}_{m} \in \mathbb{R}^N$ into a vector ${{\mathbf{x}}}'_{m} \in [0,1]^N$ by transforming each component separately, passing it through its (estimated) marginal CDF $x_n' = \hat F_{X_n}(x_n)$. By this so-called probability integral transform, we obtain a pseudo-observation vector ${\bf x}'$ whose multivariate CDF is ${\bf C}_{{\bf X}'}$. It is easy to see that each transformed random variable is uniformly distributed in $[0, 1]$. As we will see in our experimental results, transforming the original data (by applying non-linear marginal transformations) often yields improved low-rank modeling. 

\section{Algorithmic approach}
 We now turn our attention to formulating the constrained optimization problem and the algorithmic approach for low-rank fitting of the grid-sampled CDF tensor. Given a dataset $\mathcal{D}$, one can estimate and instantiate (a discretized version of) the empirical joint CDF tensor $\widehat{\mathcal{F}}$ using (\ref{empirical}). We consider a rank-$R$ approximation ${\mathcal{F}}$, computed using squared loss as the fitting criterion and solve the following optimization problem 
\begin{align} 
{ \displaystyle \min_{\pmb{\lambda}, \mathbf{A}_{1},\ldots,\mathbf{A}_{N}}\displaystyle } & 
{ \left \| \widehat{\mathcal{F}} - [\![\pmb{\lambda}, \mathbf{A}_1,\ldots,\mathbf{A}_{N} ]\!]\right  \|_F^2 } \nonumber\\ { \text{subject to} } \ & { \quad  \boldsymbol{\lambda}\geq \mathbf{0}, {\mathbf{1}^{T}\pmb{\lambda} = 1, } } \nonumber\\ &{\quad { \mathbf{A}_n(0,h)\geq0, \mathbf{A}_n(I_n,h)=1}, } \nonumber\\  & {  \quad { \mathbf{A}_n(i_n,h)\leq\mathbf{A}_n(i_n+1,h),} }\nonumber\\ &{\quad  n\in[N], \forall h\in[R]. } 
\label{optim}
\end{align} 
Although the problem described in ($\ref{optim}$) is very challenging, accurate decompositions can be efficiently computed in many practical settings. For smaller $N$, a straightforward way to estimate the latent factors is to directly instantiate tensor $\widehat{\mathcal{F}}$ and decompose it by employing alternating optimization (AO). Each model parameter is cyclically updated while the remaining ones  are fixed at their last updated values. Then, the optimization problem with respect to each $\mathbf{A}_n$ reduces to a least-squares problem
under ``valid CDF'' constraints on its columns. To solve these two sub-problems, we propose the Alternating Direction Method of Multipliers (ADMM) algorithm \cite{huang2016flexible} because it amortizes certain expensive operations and uses a warm start to handle the constraints more easily. We refer to the overall approach as AO-ADMM.

However, when the problem is high-dimensional, computing and directly forming the full grid-sampled empirical CDF tensor could be very expensive, even intractable computationally and memory-wise. Fortunately, we can still optimize the proposed model parameters without explicitly forming the empirical CDF tensor, by minimizing $\left ( \frac{1}{M}\sum_{m=1}^M \mathbf{1}\left({{\mathbf{x}}}_m\leq{{\mathbf{x}}_{\boldsymbol{i}}}\right) - \left  ( \circledast_{n=1}^N (\mathbf{A}_n(i_n,:)) \right) \boldsymbol{\lambda} \right )^2.$

Each parameter can be estimated on-the-fly by applying stochastic approximation -- i.e., sampling parts of the data at random and using the sampled piece to update the latent factors. Using this observation, we randomly sample a subset of data realizations from the training set and update the pertinent parts of the latent factors of the CPD model from rank-1 measurements of the ``hidden'' CDF tensor, using the sampled entries of the tensor. In the case of continuous variables, the conditional PDFs can be estimated from their corresponding CDF samples stored in $\mathbf{A}_n$ by interpolation and $1-D$ differentiation. For terminating the algorithm we compute the MSE on a validation set and stop if the number of maximum iterations has been reached or the MSE has not improved in the last $T$ iterations. Denoting the parameter set $\boldsymbol{\lambda},\{\mathbf{A}_n \}_{n=1}^D$ as $\vtheta$, the full procedure is shown in Algorithm~\ref{alg:FSA_HTF_SGD}. This algorithm uses the Adam optimizer with learning rate set to $0.01$.
 
\begin{algorithm}
\caption{CDF-CPD (Projected Adam)}
   \label{alg:FSA_HTF_SGD}
\begin{algorithmic}
   \STATE {\bfseries Input:} Raw data or  transformed data $\{{\mathbf{x}}_{m}\}_{m=1}^M$ in $\{\mathbf{X}$, $\mathbf{X}_{\rm val}\}$, $R$, $I$, initial learning rate $\alpha$, ${M}_{\rm batch}$
   \STATE {{\bfseries Output:} {Model parameters $\vtheta^*$}}
   \STATE Initialize model parameters $\vtheta_0$
   \REPEAT
   \STATE Sample ${M}_{\rm batch}$ data points
\State {Compute empirical estimates via Equation (\ref{empirical})}
\State Jointly update model parameters 
$\vtheta \leftarrow \vtheta -\alpha \nabla_{\vtheta} \mathcal{L}$
\State Project each $h-$th column of $\mathbf{A}_n$ using   $\mathcal{P}_{\mathcal{C}}\left ( \mathbf{A}_n(,h)\right)$
\State Project $\boldsymbol{\lambda}$  onto the probability simplex 
   \STATE Compute ${\rm MSE}_{\rm val}$ using  $\mathbf{X}_{\rm val}$
   \UNTIL{$ \rm{max_{iter}}$ is reached or ${\rm MSE}_{\rm val}$ stops diminishing }
\end{algorithmic}
\end{algorithm}

\begin{table*}[h!]
\bgroup
\footnotesize
\begin{center}
\caption{Average log-likelihood comparison in four UCI datasets. Best performances are in bold.}
\label{tbl:densityestimation}
\begin{tabular}[c]{l c c c c }
{Methods}

& {Power}& {Gas} & {Hepmass} &{Miniboone}\\
\hline
RealNVP \cite{Dinh2017NVP} & $0.17 \pm{0.01}$ & $8.33\pm{0.14}$ & $-18.71\pm{0.02}$ & $-13.55 \pm{0.49}$ \\
Glow \cite{Kingma2018GLOW} & $0.17\pm{0.01}$ & $8.15\pm{0.40}$ & $-18.92 \pm{0.08}$ & $-11.35 \pm{0.07}$ \\
MADE MoG \cite{germain2015made} & $0.40 \pm{0.01}$ & $8.47 \pm{0.02}$ & $-15.15 \pm{0.02}$ & $-12.27\pm{0.47}$\\
MAF-affine MoG \cite{papamakarios2017masked} & $0.24\pm{0.01}$ & $10.08\pm{0.02}$ & $-17.73 \pm{0.02}$& $-12.24\pm{0.45}$ \\
MAF-affine w/o MoG \cite{papamakarios2017masked} & $0.30\pm{0.01}$ & $9.59\pm{0.02}$ & $-17.39\pm{.02}$ & $-11.68\pm{0.44}$\\
FFJORD \cite{grathwohl2019ffjord} & $0.46 \pm{0.01}$ & $8.59\pm{0.12}$ & $-14.92 \pm{0.08}$& $-10.43\pm{0.04}$ \\
NAF-DDSF \cite{Huang2018NAF} & $0.62\pm{0.01}$ & $11.96\pm{0.33}$ & $-15.09\pm{0.40}$ & $-8.86 \pm{0.15}$ \\
LRCF - (HTF) \cite{amiridi2021low} & $0.68\pm0{.01}$ & $\textbf{12.15}\pm{0.02}$ & $-13.78\pm{0.02}$ & $-11.08\pm{0.04}$ \\
LR-CDF & $0.91\pm{0.01}$ & $\textbf{12.15}\pm{0.08}$ & $-11.71\pm{0.32}$ & $-7.97\pm{0.07}$ \\
{LR-Copula} & $\textbf{0.97}\pm{0.01}$ & $9.73\pm{0.05}$ & $\textbf{-9.3}\pm{0.18}$ & $\textbf{-7.94}\pm{0.08}$  \\
 \bottomrule
	\end{tabular}
	\label{tab:uci1}
\end{center}
\egroup
\end{table*}  
In Algorithm~\ref{alg:FSA_HTF_SGD}, $\mathcal{P}_{\mathcal{C}}$ is the projection operator onto a set that defines a ``valid CDF''. We project each conditional by imposing the following constraints on each factor matrix: extreme point values: $\mathbf{A}_n(I_n,h)=1$, positivity, and non decreasing $\mathbf{A}_n(:,h)$ for each $h$. For the latter, isotonic regression \cite{barlow1972isotonic} finds the best least squares fit according to the following process: When a violation $\mathbf{A}_n(i_n,h)>\mathbf{A}_n(i_n+1,h)$ is encountered, we replace this pair by their average, and back-average to the previous value as needed, to get monotonicity. We continue this process, until finally we reach $\mathbf{A}_n(I,h)$.  Each factor matrix is initialized by random non-negative initialization, followed by sorting.

\section{Experimental Study}
In this section, we showcase the effectiveness of the proposed methods, namely, LR-CDF and LR-Copula (the main difference between the two methods is the use of preprocessing in the LR-Copula case), and the proposed fitting algorithms, i.e., AO-ADMM and projected Adam, against relevant baselines in a variety of machine learning tasks.

\textbf{Low-Dimensional Toy Experiments:} We first apply our LR-CDF (AO-ADMM) model to a range of synthetic datasets to test LR-CDF’s ability to learn a known, ground truth distribution. In the $2$D case, six different toy datasets have been considered:  8gaussians, circles, moons, pinwheel, swissroll, and checkerboard. We train the proposed model on $M=5000$ samples using $I_1=I_2=30$, and $R=10$. For qualitative evaluation, we visualize (see Figure \ref{fig:toy} in supplementary material Section \ref{3dcase}) the heatmaps of the generated samples. With an appropriate choice of parameters, the proposed model is precise at capturing the modes and structure of all the datasets considered. 

In the $3$D case, we assume that a three dimensional PDF is a mixture of two Gaussian distributions with means 
$\mu_1=[-1.7,0.45,-2.50],\mu_2= [1.8,2.30,-1.20]$ and co-variance matrices $\Sigma_1={\rm diag}([{0.8}^2, {1.4}^2, {0.7}^2])$ and $\Sigma_2={\rm diag}([{1.3}^2,  {0.8}^2, {0.8}^2 ])$. We show (see the Figure in supplementary material Section \ref{3dcase}) that given $5000$ samples, the proposed method can reveal (samples of) each marginal CDF -- in this case $\boldsymbol{\lambda}$ holds the probability of choosing component.  Then, given sufficient number of CDF samples, linear interpolation and  differentiation enable high-quality conditional PDF reconstruction. Using only $I=20$ transition levels and true rank $R=2$, the recovered PDF essentially coincides with the true PDF.

\begin{table*}
\footnotesize

	\begin{center}
		\caption{Mean-squared error for single imputation for various UCI data sets (mean and standard deviations).}
	\begin{tabular}{llllllll} 

		& \emph{Banknote} & \emph{Breast}& \emph{Concrete} & \emph{Red} & \emph{White} & \emph{Yeast} \\ \midrule
				Mean                 &$1.020 \pm0.032$ &$1.000 \pm0.047$  &$ 1.010 \pm0.035$&$ 1.000 \pm0.030$&$1.000 \pm0.020$&$1.060 \pm0.052$\\
				
		Copula-EM         &$ 0.604 \pm0.040$ &$ 0.298 \pm0.035$  &$0.529 \pm0.18$&$ 0.635 \pm0.072$&$0.772 \pm0.021$ &$ 0.998 \pm0.021$\\
		
				MIWAE         &$ 0.446 \pm 0.038$ 
				              &$ 0.280 \pm0.021$  
				&$0.501 \pm0.040$&
				$ 0.643 \pm0.026$& $0.735 \pm0.033$&$ 0.962 \pm0.051$\\
				
		MAF     &$ 0.662 \pm0.029$ &$ 0.749 \pm0.084$
		&$0.974 \pm0.083$& $ 0.892 \pm0.084$&$0.855 \pm0.020$ 
		
		&$ 0.950\pm0.041$\\
		
		Real-NVP                 &$ 0.728 \pm0.046$ &$ 0.831 \pm0.029$ &$ 0.933\pm0.075$&
		$ 0.972 \pm0.029$ &$0.917 \pm0.021$&$0.994 \pm0.053$\\
				
				LR-CDF         &$ 0.574 \pm0.035$ &$\mathbf{ 0.268 \pm0.018}$ &$\mathbf{0.500\pm0.016}$&	$ 0.633 \pm0.090$ &$0.743 \pm0.012$
				
				&$ 0.947 \pm0.064$\\
				
						LR-Copula    &$ \mathbf{0.439 \pm0.038}$ &$ {0.272 \pm0.018}$  &$ {0.560 \pm0.042}$&$\mathbf{0.622 \pm0.021}$&$\mathbf{0.711 \pm0.032}$
						
						&$ \mathbf{0.934\pm0.021}$ \\
 \bottomrule
	\end{tabular}
		\end{center}

	\label{tab:uci}
\end{table*}

\begin{table*}
\begin{center}
\footnotesize
\caption{Misclassification error on different UCI datasets.}
\begin{tabular}[c]{l c c c c }
{}
& {Income}& {Credit} & {Heart} &{Car} \\
\hline
Naive Bayes   & $0.206\pm0.005$ & $0.140\pm0.018$ & $0.166\pm0.026$ & $0.152\pm0.017$\\
SVM   & $0.179\pm0.004$ &
$0.146\pm0.02$ &
$0.170\pm0.053$ &
$0.151\pm0.016$\\
LR-PMF &$0.175\pm0.003$ & $0.129\pm0.018$ & $0.147\pm0.023$ & $0.089\pm0.015$  \\
LR-CDF  & $\mathbf{{0.164\pm0.004}}$ & $\mathbf{0.102\pm0.016}$ & $\mathbf{0.122\pm0.026}$ & $\mathbf{0.069\pm0.008}$\\
\toprule
\end{tabular}
\label{densityestimation}
\end{center}
\end{table*}
\begin{table}
\begin{center}
\footnotesize
    \begin{tabular}{l c }
                        & Exchange Rates                            \\
        \midrule
        Gaussian           & $0.927 \pm 0.102 $            \\ 
        Vine             & $0.281 \pm 0.037 $            \\
        Vine-TLL2       & ${0.255}\pm 0.049 $     \\
        LR-Copula (ours)      & $\mathbf{0.172 \pm 0.108} $            \\
        \bottomrule
    \end{tabular}
        \caption{MSE and standard deviation for Exchange Rates  (lower is better).}
    \label{table:real_data_results}
    \end{center}
\end{table}
\textbf{Density estimation results:} In this section, we show results on density estimation for the following multivariate datasets: Power, Gas, Hepmass, and Miniboone dataset. Because Power ($N=6$) and Gas $(N=8)$ have lower dimensionality, we employ AO-ADMM for fitting. On the other hand, for Hepmass and Miniboone,  we use projected Adam. For training our model, we estimate the rank $R$ and the number of cutoffs $I$, considering $I_1=I_2=\cdots=I_N=I$ for simplicity. We pick the pair that best fits the data by reserving $20\%$ of the training set for validation. The set from which the tensor rank is selected is $R=[10,20,30,50,80,100]$ and the set from which we pick the number of levels is $I=[10,20,30,50]$. In Table \ref{tbl:densityestimation}, we show average log-likelihoods results over test sets, which comprise test log-likelihoods and error bars of $5$ standard deviations.

 We compare the results of the two models introduced in this work (LR-CDF and LR-Copula) with state-of-the-art methods for density estimation.  The baselines considered here are RealNVP~\citep{Dinh2017NVP}, Glow~\citep{Kingma2018GLOW}, MADE MoG~\citep{germain2015made}, MAF-affine with and without MoG~\citep{papamakarios2017masked}, FFJORD~\citep{grathwohl2019ffjord}, NAF-DDSF~\citep{Huang2018NAF}, and LRCF -(HTF) \cite{amiridi2021low}. Both of our methods significantly exceed the performance of state of the art models, suggesting an improved true density function approximation, for the
Power, Hepmass, and Miniboone datasets. We stress that even though our method aims for CDF estimation, it performs better in PDF (density) estimation as well, compared to state of art density estimation methods. This is quite surprising on first sight, and it is in part due to the use of far-away samples to estimate the density in sample-starved areas, which is built-in the empirical CDF. The same holds for the characteristic function approach in \cite{amiridi2021low}, but still the proposed CDF approaches are better -- which speaks for the appropriateness of the proposed formulation.

\textbf{Imputation results:} 
Next, we focus on the \emph{imputation problem}: given a trained joint distribution model, and an incomplete realization $\mathbf{x}=[\mathbf{x}_\text{OBS}, \mathbf{x}_\text{MISS}]^T$, our next task is to infer the missing entries directly. In our experiments, all six datasets (we only consider continuous features) from the UCI database considered, are corrupted by removing $30\%$ of the features uniformly at random. Our models are fitted with AO-ADMM for Banknote, Concrete, and Yeast and projected-Adam for the rest. The experiment compares the MSE of the proposed methods with other well established baselines. We chose the mean imputation, which is the simplest baseline, the Copula-EM approach \cite{zhao2020missing}; the recently proposed MIWAE approach \cite{mattei2019miwae}; MAF, and Real-NVP. The results, averaged over 5 randomly corrupted repetitions, are presented in Table \ref{tab:uci1}. The results suggest that both approaches provide more accurate imputations than all competitors, with CPD-Copula outperforming the rest in most cases.

\textbf{{PMF or CDF modeling?}}
We also present an important set of experiments which targets to answer the following question: Which CPD model (LR-PMF~\cite{kargas2018tensors} or LR-CDF) offers better performance on modeling mixed data? We test our LR-CDF (Adam) model using four different datasets (of mixed variables) for classification against the CPD-based joint PMF model~\cite{kargas2018tensors} and two other standard baselines, namely, Naive Bayes classifier and linear SVMs. The results on Table~\ref{densityestimation} suggests that modeling the joint CDF instead of the joint PMF, as proposed in ~\cite{kargas2018tensors}, may achieve a better performance in misclassification error. This further supports the merits of the proposed method, which can be useful even for discrete variables, for which we naturally turn to the PMF.  

\textbf{A Copula comparison:} Modeling with Copulas is a popular tool in financial risk management as they are invariant to nonlinear feature scaling and useful for modeling extreme tail distributions. In the following experiment, we consider a data set of size $N = 5844$ that contains $15$ years of exchange rates of Yen, Euro, Canadian dollars and the British Pound to US dollars. We estimate a parametric Gaussian copula, a Vine copula \cite{aas2009pair}, a Vine copula with only TLL2 pair-copulas \cite{nagler2020pyvinecopulib}, and our CPD LR-Copula model (via AO-ADMM) for this dataset, and then perform data imputation, as described earlier. The proposed LR-Copula model yields significantly smaller MSE relative to the baselines, which we attribute to the unassuming non-parametric nature and strong identifiability properties of our LR-Copula model.



\section{Conclusions}
In this paper, we introduced a new class of models for general purpose distribution estimation. Our work has shown that {\em every} multivariate CDF admits a compact grid-sampled approximation in terms of a latent variable naive Bayes model with a bounded number of hidden states through CPD corresponding to the tensor rank. We have considered the statistical and computational efficiency of the proposed estimators and highlighted desired properties such as easy marginalizability over subsets of variables, fast sampling, and identifiability under mild assumptions. Furthermore, our experimental results indicate that estimating a grid-sampled CDF yields better data modeling than direct PDF estimation. We have also presented results that clearly support the broad applicability of low-rank grid-sampled CDF estimators in various tasks of interest in machine learning and statistics.

\section{Supplementary Material}
\noindent In this section, we provide a proof for sample complexity analysis of grid-sampled CDF-CPD estimators and additional experimental material to better evaluate the performance of the proposed method.
\subsection{Sample Complexity}
\label{sample_complx}
Using Lemma C.1. of \cite{daskalakis2012learning}, we have the following sample complexity result for our problem.

\textbf{Lemma 1:} \textit{ Let $F$ be any unknown $N-$ dimensional (``true'') grid sampled CDF of ${\mathbf{X}}$ with rank $R$, alphabet size $I$ (per variable), and resolution parameter $\epsilon'$. There is an algorithm that uses 
\resizebox{.9\linewidth}{!}{\
  \begin{minipage}{\linewidth}
  \begin{equation*}
\mathcal{O}\left( {\epsilon'}^{-2}({RNI\text{log}\frac{1}{\epsilon'}+NI R\text{log} R+RN^2I
\text{log}(2I)}) \cdot \text{log}(\frac{1}{\delta})\right)
\end{equation*}
  \end{minipage}}
}

\textit{samples from $F$, and with probability $1-\delta$ outputs a distribution $F'$ that satisfies $d_{TV}(F,F')\leq 6{\epsilon'}.$}

\noindent\textbf{Proof}:
\noindent Let us start by considering the 2D rank-1 case. 

We consider a dictionary of rank-1 CDFs generated by latent factor vectors whose elements are drawn from a finite quantization. An element of the grid-sampled CDF matrix $\hat{\mathbf{F}}(i,j)$ can be expressed as
\begin{equation*}
\hat{\mathbf{F}}(i,j)=(a_i+\tilde{\epsilon}_i)(b_j+\tilde{\epsilon}_j)\end{equation*}
\noindent where $a_i$ and $b_j$ are the exact latent factors of ${\mathbf{F}}(i,j)=a_i b_j$. If the quantization error for both dimensions is no more than $\epsilon$, i.e., $|\tilde{\epsilon}_i|\leq\epsilon$, $|\tilde{\epsilon}_j|\leq\epsilon$, and using that the factors are 1-D CDFs (follows from marginalization) and thus $a_i\leq1, b_j\leq1$, there exists a rank-1 CDF in the dictionary such that
\begin{equation*}
|{\mathbf{F}}(i,j)-\hat{\mathbf{F}}(i,j)|\leq2\epsilon+{\epsilon}^2,\end{equation*}
 and thus, the overall error becomes
 \begin{equation*}
 d_{TV}(\mathbf{F}, \hat{\mathbf{F}})\leq IJ (2\epsilon+{\epsilon}^2)
\end{equation*}

\noindent Assuming that $I_1=I_2=\cdots=I_N=I$ for simplicity, the bound in the $N$-dimensional rank-1 case is  
 \begin{equation*}
d_{TV}(\mathbf{F}, \hat{\mathbf{F}}_\text{rank-1})\leq {(2I)}^N\epsilon.
\end{equation*}
This is because there are $I^N$ grid-sampled CDF elements overall, and the expression for the per-element error involves a total of $2^N-1$ terms (product of $N$ pairwise sums, with one term corresponding to the exact value of the CDF), each of which is bounded above by $\epsilon < 1$. 
In the $N$-dimensional case of rank-$R$, the overall error is bounded by 
 \begin{equation*}
d_{TV}(\mathbf{F}, \hat{\mathbf{F}}_\text{rank-$R$})\leq {R(2I)}^N\epsilon.
\end{equation*}

 We will denote this bound as $\epsilon'$, $\epsilon'={R(2I)}^N\epsilon$. 
 
 By choosing $b$ bits to quantize each element of a conditional CDF, the resolution that we get is $\epsilon=\frac{1}{2^b}$. The number of distributions in our quantized dictionary is then $M=2^{RNIb}$.

\noindent By focusing on $M$, 
\begin{equation*}
    M=2^{RNIb}=2^{-{RNIlog{{\epsilon}}}}
\end{equation*}
  and replacing  
  \begin{equation*}
    \epsilon=\frac{\epsilon'}{{R(2I)}^N},
\end{equation*}
 we get that 
\begin{align*}
    M&=2^{RNIb}\\
    &=2^{-RNI(log\epsilon'-log{({{R(2I)}^N})})}\\
    &=2^{RNIlog\frac{1}{\epsilon'}+RNI(logR+log{(2I)}^N)}\\
    &= 2^{RNIlog\frac{1}{\epsilon'}+NI RlogR+RN^2Ilog(2I)}.
\end{align*}

\noindent By plugging in $M$ in Lemma C.1 considered in {\cite{daskalakis2012learning}} and replacing the worst case number of samples $ \mathcal{O}({\epsilon'}^{-2}logM \cdot log(\frac{1}{\delta})),$ we get the final result,

\resizebox{.9\linewidth}{!}{\
  \begin{minipage}{\linewidth}
  \begin{equation*}
\mathcal{O}\left( {\epsilon'}^{-2}({RNI\text{log}\frac{1}{\epsilon'}+NI R\text{log} R+RN^2I
\text{log}(2I)}) \cdot \text{log}(\frac{1}{\delta})\right).
\end{equation*}
  \end{minipage}
}

\subsection{2D, 3D-Synthetic Experiments}\label{3dcase}
 \begin{figure*}
\begin{center}
    \includegraphics[width=6.0in]{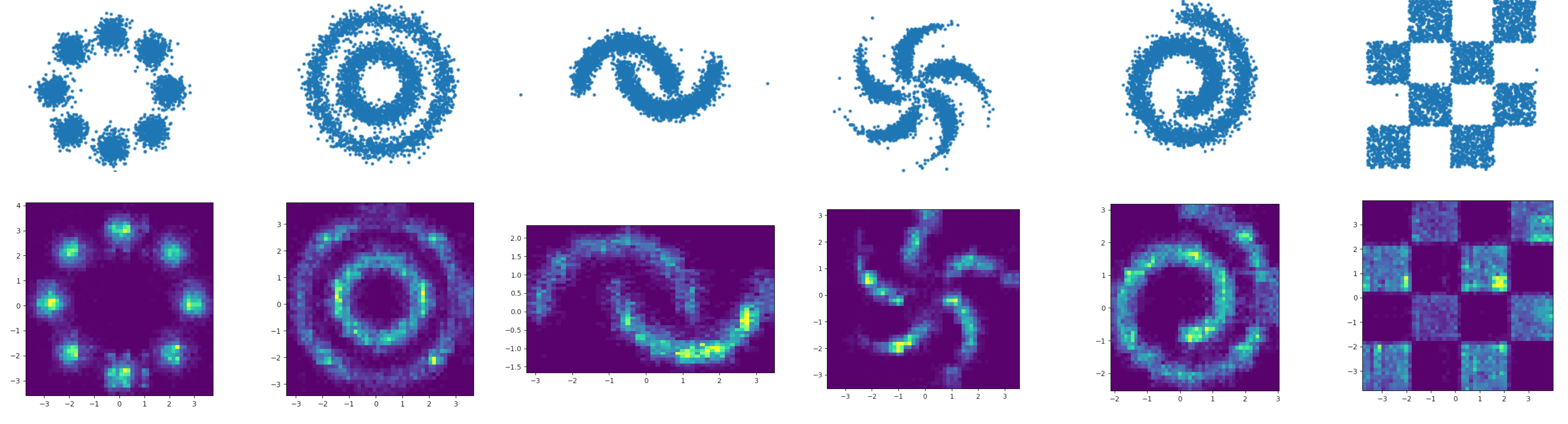}  
            \caption{Visualization of training samples (upper row) and synthetic samples generated with a learned CPD-CDF (lower row).} 
   \label{fig:toy} 
\end{center}
\end{figure*}

\begin{figure*}
\begin{center}
{\includegraphics[width=0.23\linewidth]{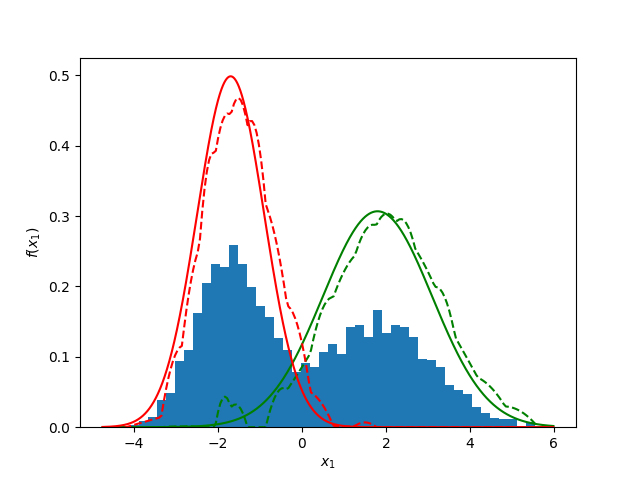}} \quad
{\includegraphics[width=0.23\linewidth]{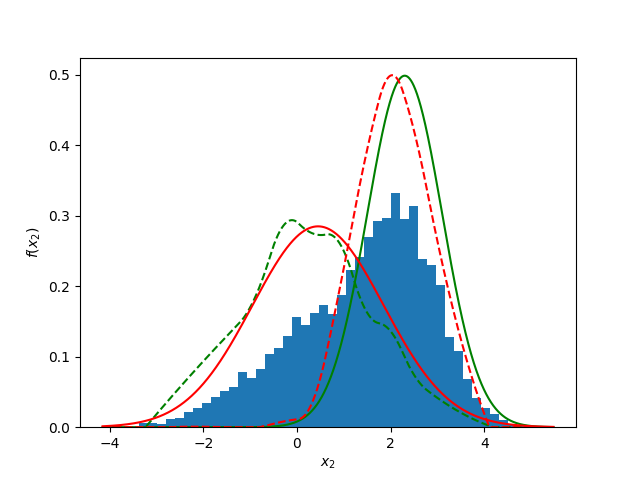}} \quad
{\includegraphics[width=0.23\linewidth]{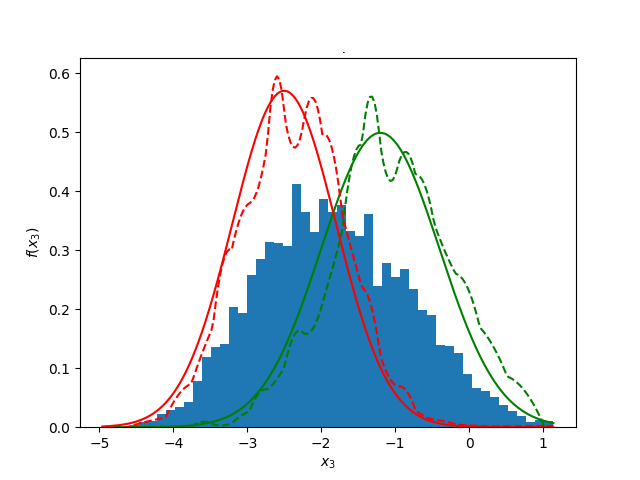}}

{\includegraphics[width=0.23\linewidth]{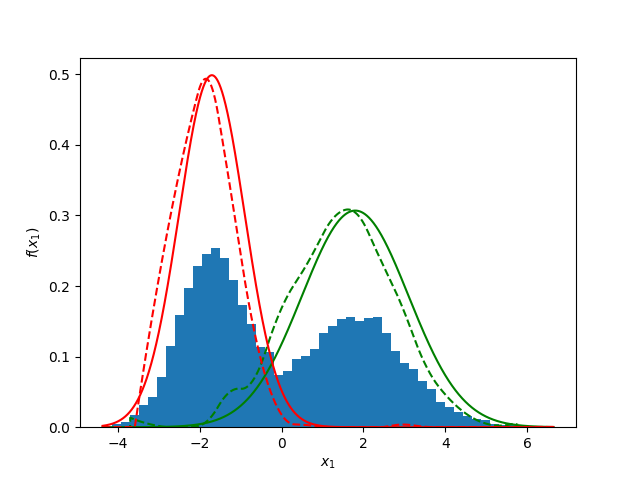}} \quad
{\includegraphics[width=0.23\linewidth]{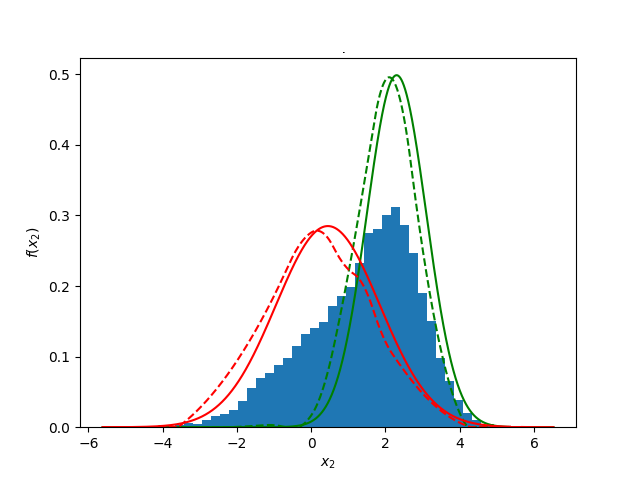}} \quad
{\includegraphics[width=0.23\linewidth]{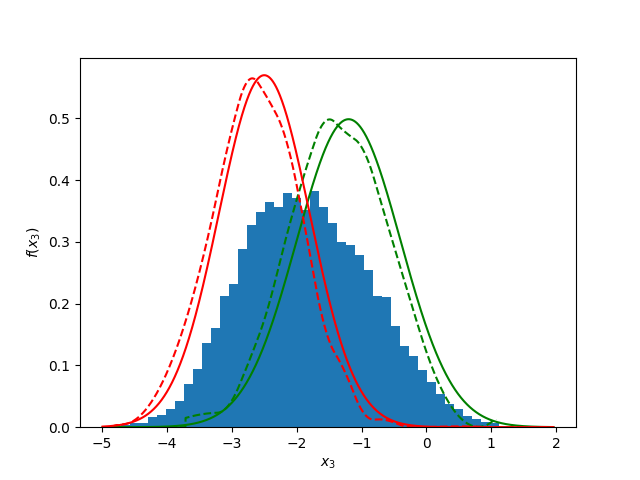}}
\caption{ Illustration of the key idea on a $3$-D two-component Gaussian mixture. Each row shows the estimated component PDFs of $X_1, X_2, X_3$ in dashed line, and the true component PDFs in solid line. We also visualize the compound histogram per dimension, given $M$ training samples. The first row corresponds to $M=2000$, $R=2$, $I=20$ and the second to $M=4000$, $R=2$, $I=20$.}
\label{marginal_gauss}
\end{center}
\end{figure*}

In this section, we employ synthetic experiments to showcase the effectiveness of the proposed algorithm. We first showcase $2$D sampling results from six different toy datasets: 8gaussians, circles, moons, pinwheel, swissroll, and checkerboard (see Figure~\ref{fig:toy}). Please refer to the discussion of implementation details in the main paper.

Next, we assume that a $3-$dimensional PDF of a random vector ${\boldsymbol X}:= \left[X_1, X_2, X_3\right]^T$ is a mixture of two Gaussian distributions with means $\mu_1=[-1.7,0.45,-2.50], \mu_2= [1.8,2.30,-1.20]$ and co-variance matrices
\[ \Sigma_1 =
  \bracketMatrixstack{
    {0.8}^2 & 0 & 0\\
    0 & {1.4}^2 & 0 \\
    0 & 0 & {0.7}^2
  }, \Sigma_2 =
  \bracketMatrixstack{
    {1.3}^2 & 0 & 0\\
    0 & {0.8}^2 & 0 \\
    0 & 0 & {0.8}^2 
  }.
\] We show (see Figure~\ref{marginal_gauss}) that given only $2000$ (top row) or $4000$ (bottom row) samples, the proposed method can effectively estimate samples of each marginal CDF, which after interpolation and  differentiation yield high-quality conditional PDF reconstruction. For these plots, $I=20$, $R=2$, and cubic interpolation of the marginal CDFs was used.

\clearpage
\bibliography{AIStats22.bib}

\end{document}